\documentclass{article}
\usepackage{ijcai25}
\usepackage{times}
\usepackage{soul}
\usepackage{url}
\usepackage[hidelinks]{hyperref}
\usepackage[utf8]{inputenc}
\usepackage[small]{caption}
\usepackage{graphicx}
\usepackage{amsmath}
\usepackage{amsthm}
\usepackage{booktabs}
\usepackage[switch]{lineno}

\usepackage{amssymb}
\usepackage{microtype}
\usepackage{multirow}
\usepackage{subcaption}
\usepackage{algpseudocode}
\usepackage{algorithm2e}
\usepackage{xcolor}
\usepackage{mdframed}
\usepackage{makecell}
\usepackage{inconsolata}
\usepackage{tabularx}

\title{REAL: Response Embedding-based Alignment for LLMs}

\author{
Honggen Zhang$^1$
\and
Xufeng Zhao$^2$\and
Igor Molybog$^{1}$\And
June Zhang$^1$\\
\affiliations
$^1$Univerity of Hawaii at Manoa\\
$^2$University of Hamburg\\
\emails
\{honggen, molybog,zjz\}@hawaii.edu,
xufeng.zhao@uni-hamburg.de
}

\begin{document}

\maketitle

\begin{abstract}
Aligning large language models (LLMs) to human preferences is a crucial step in building helpful and safe AI tools, which usually involve training on supervised datasets. Popular algorithms such as Direct Preference Optimization (DPO) rely on pairs of AI-generated responses ranked according to human annotation. 
The response pair annotation process might bring human bias. Building a correct preference dataset is the costly part of the alignment pipeline.  To improve annotation efficiency and quality in the LLMs alignment, we propose REAL:\textbf{R}esponse \textbf{E}mbedding-based \textbf{A}lignment for \textbf{L}LMs, a strategy for constructing a high-quality training dataset that focuses on acquiring the less ambiguous preference pairs for labeling out of a set of response candidates. 
Our selection process is based on the similarity of embedding responses independently of prompts, which guarantees the selection process in an off-policy setting, avoiding adaptively measuring the similarity during the training.
Experimental results on real-world dataset SHP2 and synthetic HH-RLHF benchmarks indicate that choosing dissimilar response pairs enhances the direct alignment of LLMs while reducing inherited labeling errors. The model aligned with dissimilar response pairs obtained a better margin and win rate on the dialogue task. Our findings suggest that focusing on distinct pairs can reduce the label error and improve LLM alignment efficiency, saving up to $65\%$ of annotators' work. The code of the work can be found \url{https://github.com/honggen-zhang/REAL-Alignment}.
\end{abstract}

\section{Introduction}
Large Language Models (LLMs), empowered by the enormous pre-trained dataset from the Internet, show the power to generate answers to various questions and solutions to challenging tasks. However, they might generate undesirable content that is useless or even harmful to humans \cite{wang2024essence,ouyang2022training}.
Additional training steps are required to optimize LLMs and align their responses with human preferences. For that purpose, Reinforcement learning from human feedback (RLHF) \cite{christiano2017deep,ouyang2022training}  is proposed, which consists of
estimating the human preference reward model from response preference data and steering LLM parameters using a popular reinforcement learning algorithm of proximal policy optimization~\cite{schulman2017proximal}. RLHF requires extensive computational resources and is prone to training instabilities. 

Recently, direct alignment from the preference approach, which does not explicitly learn the reward model, has emerged as an alternative to RLHF \cite{zhao2022calibrating,song2023preference,zhao2023slic,xu2023contrastive,rafailov2024direct}. Direct Preference Optimization (DPO) \cite{rafailov2024direct,azar2023general} is a milestone method. It formulates the problem of learning human preferences through finetuning LLM with implicit reward model using a static dataset $\mathcal{D} = \{y_{i}^+, y_{i}^-, x_{i}\}_{i=1}^N$, where $x_i$ is the $i$th prompt, $ y_{i}^+, y_{i}^-$ are the corresponding preferred and rejected responses. DPO requires the explicit preference signal from an annotator to label the response pair. However, constructing high-quality $\mathcal{D}$, which highly agrees with human preference, is a challenge. 
Existing human labeled $\mathcal{D}$ has incorrect and ambiguous preferences pairs (30\%-40\%)~\cite{bai2022training,ouyang2022training,stiennon2020learning,wang2024secrets}. 
 Some DPO variations, such as Contrastive Post-training \cite{xu2023contrastive}, RPO~\cite{song2023preference} were proposed to augment $\mathcal{D}$ using AI as labelers but might generate low-quality pairs. 
Some other DPO variation methods\cite{guo2024direct,yu2024direct} actively choose better samples using additional annotators at the cost of increased computations. 
Unlike such methods to use AI as an additional labeler to augment the dataset or adaptively select, we construct a high-quality preference dataset without adding extra annotation computation resources in the off-policy setting.

\begin{figure*}[t]
     \centering
         \includegraphics[width=1\textwidth]{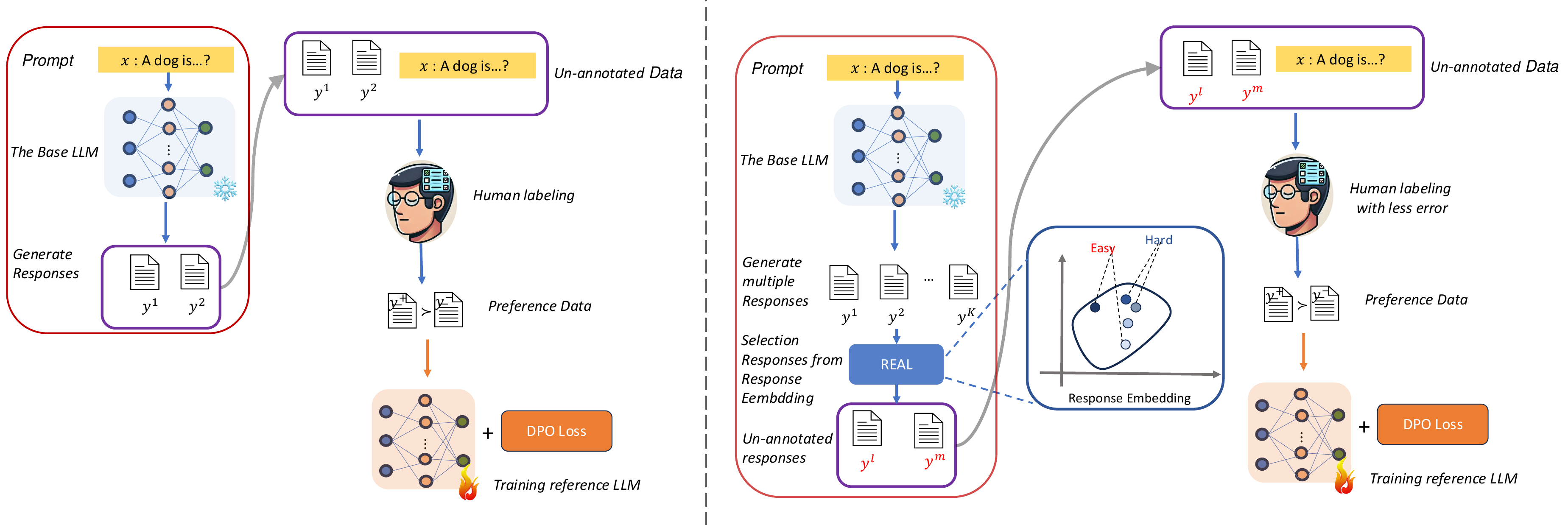}
         \caption{Comparison between vanilla DPO and REAL. Left: The DPO data construction and training process. Right: The REAL method. For each prompt, multiple responses will be generated. We extract the embeddings of the response based on the LLM to select the high-quality pair to label, which reduces ambiguity and contributes better to  DPO alignment. }
        \label{fig:diagram}
\end{figure*}

This paper proposes a novel method for enhancing DPO learning with efficient and less ambiguous preference data construction (see Fig.~\ref{fig:diagram}). We suggest constructing a valuable unannotated sub-dataset $\{y_{i}^l, y_{i}^m, x_{i}\}_{i=1}^N$ from large annotated responses $\{y^1_i,y^2_i,\cdots,y^K_i\}_{i=1}^N$ generated by AI or scripted from the Internet. 
After constructing the valuable unannotated sub-dataset with REAL, we can ask humans to label it, achieving a high-quality preference dataset $\mathcal{D} = \{y_{i}^+, y_{i}^-, x_{i}\}_{i=1}^N$ with less ambiguous labeling.  We only train DPO on the most informative subset of response pairs. 
Inspired by works in contrastive learning\cite{chen2020simple}, we connect the usefulness of response pairs to the semantic similarity (described as cosine similarity) between their representations in the embedding space. While most similar negative pairs will boost the contrastive learning to learn a better embedding for the images or sentences, it might generate more false negative pairs without accessing the complete knowledge of labeling~\cite{zhang2024hasa,chuang2020debiased}, which might hurt learning performance. On the other hand, dissimilar pairs can be preferred for the DPO owing to smaller noise in the labels under the assumption that $y^+\succ y^-$. 

However, the sentence embedding of \textbf{prompt+response} might vary during training. It will take REAL to the on-policy setting, which would result in expensive re-calculation of similarity. Regardless of the \textbf{prompt}, we observed the distribution of similarity between independent \textbf{response} keep invariant during the training, i.e. if $\text{sim}(\boldsymbol{y^+(t),y^-(t)})$, we have  $\text{sim}(\boldsymbol{y^+(t+1),y^-(t+1)})$, where the boldface is the embedding of response $y$. 

We test our method on the real-world dataset SHP2, which contains multiple responses per prompt. We want to extract a high-quality pair $(y^l_i,y^m_i)$ to label from $\{y^1_i,y^2_i,\cdots,y^K_i\}$.  In addition to considering the most similar and most dissimilar pairs of responses in the response embedding space, we implemented an approach that splits the responses into two clusters in the embedding space and selects centroids of the clusters as the centroid pairs. Given three response pairs for each prompt, the centroid pair is close to dissimilar pairs (the most dissimilar or random pair) and demonstrates the best performance according to our experimental results.
We extend our method to the HH-RLHF dataset by ranking the data tuples according to the similarity of the response pair. It shows that dissimilar pairs empirically construct the best dataset for alignment when compared on several metrics to randomly selected or similar pairs due to the higher agreement with human preference.

\begin{enumerate}
    \item We highlight the overlooked importance of sentence embeddings in LLM training: The model learning can be enhanced by investigating the sentence embeddings and integrating this information into the fine-tuning process. 
    \item  We introduce less ambiguous response pair construction strategies to acquire high-quality data without adding extract annotation effort. 
    \item  We maintain an offline dataset construction setting by the novel observation of the distribution of similarity between independent responses, keeping invariant during the training.
    \item Our experiments demonstrate that dissimilar pairs in the embedding space align better with human preferences than random or similar pairs, owing to reduced errors in the label.
\end{enumerate}

\vspace{-3mm}

\section{Background: Direct Language Model Alignment}
LLM alignment refers to training a language model to assign a higher probability to a prompt-response pair $(x, y)$ with a higher human preference reward $R_{xy}$. An estimation $r$ of the reward is used in practice. It is imperative to ensure that for a given prompt $x$, the estimated reward $r(x,y)$ is close to the true reward $R_{xy}$ for each response. The problem of learning the best estimator for the reward function can be formulated as
\begin{equation}\label{eq:reward}
    r = \arg \min_{r}\mathbb{E}_{(x,y,R_{xy}) \sim \mathcal{R}}[(r(x,y)-R_{xy})^2]
\end{equation}
where $\mathcal{R}$ is the dataset consisting of the prompt $x$, responses $y$, and oracle reward values $R_{xy}$. 

The human feedback rarely comes in the form of the true reward samples $(x, y, R_{xy})$. Since the true reward $R_{xy}$ is unknown, ranking or pairwise comparison of responses is more common to estimate the reward. In the pairwise comparison case, there are two sample responses $(y^+,y^-)$ that correspond to a single prompt $x$. Here, $y^+$ and $y^-$ are the preferred and non-preferred responses, respectively, as defined by
 human subjects providing a binary preference to label them $(r(x,y^+) > r(x,y^-)).$ This method of labeling does not explicitly require access to the true reward function $R_{xy}$.
 Alignment can still be performed by applying the RLHF algorithm using binary human preference data.  RLHF is expensive and unstable to train. Recently, methods for direct alignment from preferences have emerged to replace the RLHF when aligning the LLMs.  Direct Preference Optimization (DPO) \cite{rafailov2024direct} optimizes the policy $\pi$ directly with supervised loss, without relying on the explicit reward model and subsequently reinforcement learning. It works with a static dataset $\mathcal{D} = \{x_{i},y_{i}^+, y_{i}^-\}_{i=1}^N$ sampled from the distribution of human preferences to estimate the human preference model $p(y^+\succ y^- |x) = \sigma(r(x,y^+) -r(x,y^-)) $ the which represent by Bradley-Terry model (see appendix~\ref{appB}). 
Thus, the loss function is defined as
\begin{equation}
    \mathcal{L} = -\mathbb{E}_{x,y^+,y^-} \big[\log p(y^+\succ y^- |x) \big]
\end{equation}
In the DPO, the preference model $p(y^+\succ y^- |x)$ is directly related to the parameterized policy $\pi_{\theta}$ and reference policy $\pi_{\text{ref}}$. Instead of training an explicit reward model, DPO suggests reconstructing an implicit reward model from the aligned LLM $r = \beta \log \frac{\pi_\theta}{\pi_{\text{ref}}}$. 
\begin{align}\label{eq:dpo loss}\small
    \mathcal{L}_{\text{DPO}} = & - \mathbb{E}_{x,y^+,y^-} \big[\log \sigma \big(\beta\log\frac{\pi_{\theta}(y^+|x)}{\pi_{\text{ref}}(y^+|x)} \\-  \nonumber & \beta\log\frac{\pi_{\theta}(y^-|x)}{\pi_{\text{ref}}(y^-|x)}\big)\big],
\end{align}
where $\sigma$ is the logistic function and $\beta$ is hyper-parameter. hello
fa

In contrast to RLHF, which requires the generation of new responses for on-policy training, DPO's performance will be largely affected by the data size and annotation quality. As the size of $\mathcal{D}$ increases,  $p(y^+\succ y^- |x)$ converges to the true distribution of human preferences $p^{*}$~\cite{azar2023general}. However, obtaining the high-quality $\mathcal{D}$ from human preferences is expensive, and it is easier to produce incorrect labeling $(y^-\succ y^+)$. While some additional human relabeling and checking methods were used, there is still a gap with the general human agreement~\cite{stiennon2020learning,wang2024secrets}. Constructing high-quality human preference data is crucial for the DPO LLM alignment.

\section{Method}
We propose automated identification of high-quality response pairs from unlabeled data to reduce the error in the labeling and improve the performance of the aligned model.
Since DPO loss $\mathcal{L}_{\text{DPO}}$ (Eq.~\ref{eq:dpo loss}) is estimated based on the static empirical preference dataset $(y^+\succ y^-)$, 
this raises the question of how to build the empirical dataset $\mathcal{D}$ 
to better represent the true distribution of human preference. 

Given the prompt $x$, which is usually written by human experts, we generate a response pair $(y^l, y^m)$ and label it with manual labor or AI help. As a result, we will obtain the preference pair $(y^+,y-).$
The DPO training algorithm assumes that we have access to the preferences through a
dataset $\mathcal{D} = \{x_i,y_i^+,y_i^-\}_{i=1}^{N}$ sampled from the true human preference distribution $p^{*}$, where $N$ is the data size. 
Without knowing $p^{*}$, the human-labeled pairs might have a low agreement with the true preference of the entire human population, i.e., $p^{*}(y_i^{+}\succ y_i^{-} |x_i) < p^{*}(y_i^{-}\succ y_i^{+} |x_i)$. This label error could be caused by the preference of a particular annotator. 
The label errors affect the quality of the data, thus the alignment. 

Constructing a less ambiguous and high-quality dataset $\mathcal{D}$ matters for learning the true human preference signal without inflating the size of $\mathcal{D}$. Generating responses for a prompt using a large language model is a relatively cheap operation. Generating several responses and then picking a better-quality pair out of them may lead to a dataset $\mathcal{D}$ that better represents the true human preference distribution $p^{*}$.


\subsection{Pair Selection}\label{sec:pair_selection}

In representation learning, as the representation of extracted features, embedding similarity is commonly used to measure the data quality \cite{chen2020simple} and sentence similarity~\cite{mikolov2013efficient,zhang2024hasa}.
SLiC \cite{zhao2023slic} selects responses that are highly correlated with the golden standard $y^*,$ given by a human annotator. However, a target golden response $y^*$ need to extra effort to label and will limited the model to generate new responses better than $y^*$. 

In our dataset construction, the response pair does not depend on any golden-standard $y^*,$ making the selection process fully automatable.
In this paper, we suggest building the preference dataset $\mathcal{D} = \{y^+_i, y^-_i,x_i\}_{i=1}^N$ from the $K$-responses dataset $\{y^1_i,y^2_i,\cdots,y^K_i, x_i\}_{i=1}^N$ by selecting and labeling a single response pair per prompt $x_i$ out of ${K \choose 2}$ possible response combinations. 

To select the one pair from ${K \choose 2}$ pairs, we hypothesize that the value of a response pair for the alignment process is connected to the semantic similarity between the responses, which can be measured through their embeddings.
The selection problem faces a trade-off. On the one hand, learning from similar contrastive examples is deemed to be more useful in establishing a precise decision boundary. On the other hand, such similar responses may offer less informative learning resources, as they risk being non-representative outliers. They are also less obvious for a human to label correctly, and thus, selecting them is more likely to result in a labeling error. Our work, at its core, aims to understand the implications of this trade-off for building a high-quality dataset $\mathcal{D}.$


Encoder-based models, such as BERT~\cite{devlin2018bert}, define embeddings using the output of a special classification token (e.g., [CLS]) or by applying a pooling operation over the hidden states of all tokens. Similarly, for decoder-based LLMs, we use a pooling operation over their last hidden states to obtain embeddings~\cite{jiang2023scaling}.

Denoting the last hidden state $\boldsymbol{H} \in \mathbb{R}^{N\times L\times D}$ of the base model, where $N$ is the batch or data size, $L$ is the sequence length, and $D$ is the hidden dimension. To get an embedding of the sequences in the batch, we average $\boldsymbol{H}$ across the sequence dimension $L$. Formally, the $i$-th sequence of the batch will have a $D$ dimensional embedding
\begin{equation}
    \boldsymbol{y}_i = \frac{1}{L}\sum_{l=1}^L \boldsymbol{H}_{i,l,:}
\end{equation}
Note that we calculate the embeddings of the responses independently of their corresponding prompts ${x}_i$, i.e., the response is passed to the model without any context when embeddings are being constructed. We find empirically that this method is more suitable for achieving the goal of selecting high-quality data off-policy regardless of modeling changes (Section 4.2).
We use the common metric of normalized inner product between sequence embeddings $\frac{\boldsymbol{y}_i^T \boldsymbol{y}_j}{||\boldsymbol{y}_i||\cdot||\boldsymbol{y}_j||},$ also known as the cosine similarity $\cos(\boldsymbol{y_i},\boldsymbol{y_j} )$, to measure semantic similarity between sequences.

By applying the cosine distance in the embedding space between response pairs, we arrive at several data selection strategies.
\begin{enumerate}
    \item We define the most similar response pair as a \textbf{hard} pair, which is difficult to differentiate in the embedding space. 
    \begin{align}\label{eq:comp_hard}
    \mathcal{Y}^i_{\text{hard}}  = \{y_i^l,y_i^m,x_i\},
    \end{align}
    where, 
    \begin{equation}
        l,m = \arg \max_{j\ne k}  \cos(\boldsymbol{y}_i^j,\boldsymbol{y}_i^k)
    \end{equation}
    For $\forall {y}_i^j,{y}_i^k\in \{{y}_i^1,y_i^2,\ldots,y_i^K\}$, $j\ne k$, given the particular prompt $x_i$.
    \item We consider the most dissimilar pair to build the \textbf{easy} pair. The easy response pair are obviously be differentiated by the base LLMs. It is defined as:

    \begin{align}\label{eq:comp_easy} 
    \mathcal{Y}^i_{\text{easy}}  = \{y_i^l,y_i^m,x_i\},
    \end{align}
    where
    \begin{equation}
    l,m =   \arg \min_{j\ne k}  \cos(\boldsymbol{y}_i^j,\boldsymbol{y}_i^k),
    \end{equation}
    \item Furthermore, we consider a way to balance the benefits of selecting the most similar and the most distinct responses by developing a method to select the most representative response pair. For that, the responses $\{y_i^1, \ldots, y_i^K\}$ that correspond to a particular prompt $x_i$ are being split into two clusters $C_1$ and $C_2$ using the K-means algorithm in the space of response embeddings. Let's denote the centers of the clusters $C_1$ and $C_2$ as $\mathbf{u_1}$ and $\mathbf{u_2}.$

    We define the \textbf{centroid} pairs by selecting responses closest to the clusters' centers.
    \begin{equation}\label{eq:centroid}
    \mathcal{Y}_{\text{centroids}} = \{y^l_i, y^m_i,x_i\}, 
    \end{equation}
    where
    \begin{align}
        \nonumber y^l_i = \arg \min_{y^l_i \in C_1}||\boldsymbol{y}^l_i - \boldsymbol{u}_1||^2\\ \small\nonumber
        y^m_i = \arg \min_{y^m_i \in C_2}||\boldsymbol{y}^m_i - \boldsymbol{u}_2||^2,
    \end{align}
    Note that once the embeddings are normalized, minimizing the Euclidean distance $||\boldsymbol{y}_i-\boldsymbol{u}_1||^2$ is equivalent to maximizing the cosine similarity $\cos(\boldsymbol{y}_i,\boldsymbol{u}_1)$. 
\end{enumerate}

We also build the \textbf{random} baseline $\mathcal{Y}^i_{\text{random}}$ which is sampled from the same distribution as the standard recipe suggests (a single response pair per prompt) and remains free from biases introduced by prior knowledge. To do that, we select $\mathcal{Y}^i_{\text{random}} = (y^1_i,y^2_i)$ uniformly from the ${K \choose 2}$ possible response pairs.

 After we automatically selected the particular pair from ${K \choose 2}$ possible response combinations, human effort will be paid to label them. The human labeler will choose the preference $y_i^+$ and the non-preference $y_i^-$ based on the given $x_i$ given $\{y^l_i, y^m_i,x_i\}$. Thus, we will obtain alignment sub-datasets $\mathcal D$: $\mathcal{D}_{\text{hard}}$, $\mathcal{D}_{\text{easy}}$, $\mathcal{D}_{\text{centroids}}$, and $\mathcal{D}_{\text{random}}$. We use them within the DPO procedure to align LLMs.  
 


\subsection{Consistent Responses Embedding Over Training}

DPO-based LLMs alignment method provides the benefit of avoiding the reward model and generating on-policy responses when compared to RLHF. The off-policy setting will make the training process effective. While the embedding space might vary during the training, to keep the off-policy setting, we investigate the distribution of cosine similarity during the DPO finetuning process. 

\begin{figure}[t]
     \centering
     \begin{subfigure}[t]{0.45\textwidth}
         \centering
         \includegraphics[width=\textwidth]{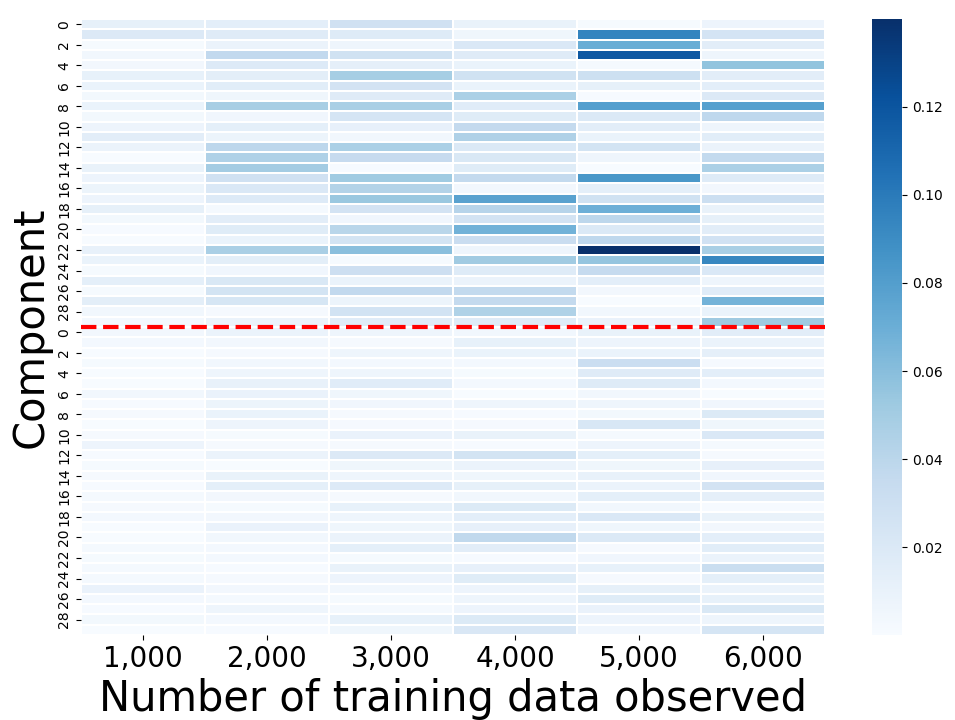}
         \caption{The first 20 components of the embedding of the joint string $\boldsymbol{[x,y]}$ (top) and the first 20 components of the embedding $\boldsymbol{y}$ (bottom), obtained using 6 different DPO checkpoints.}
         \label{fig:embed diff}
     \end{subfigure}
     \begin{subfigure}[t]{0.45\textwidth}
         \centering
         \includegraphics[width=\textwidth]{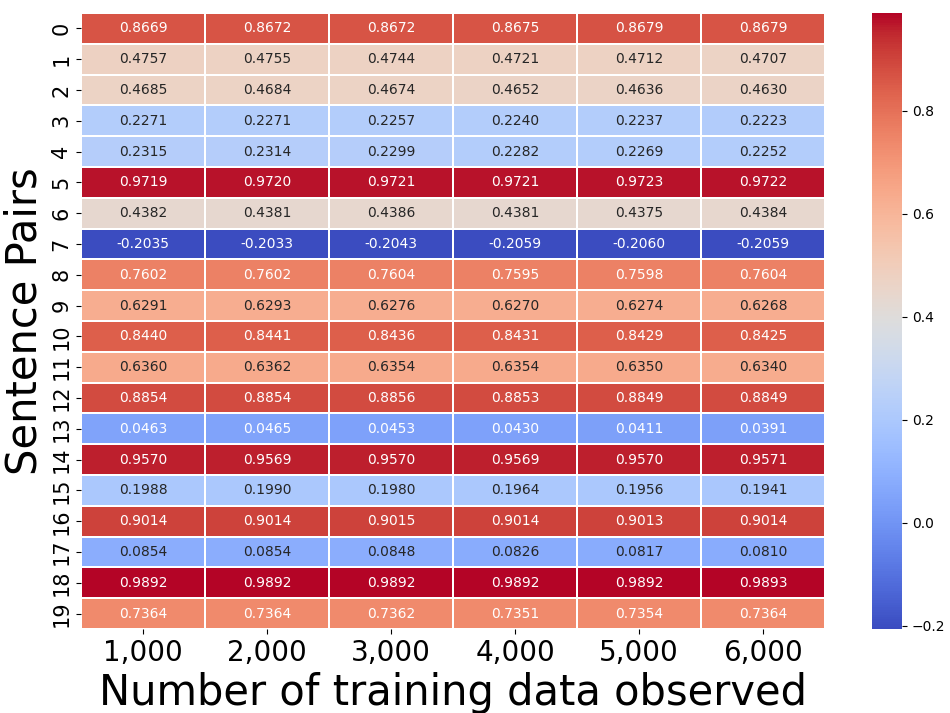}
         \caption{The response pair similarity $\cos (\boldsymbol{y}^l_i, \boldsymbol{y}_i^m)$ changes over the DPO training process (20 pairs of responses on 6 checkpoints)}
         \label{fig:embed pair}
     \end{subfigure}
        \caption{Sentence embedding and similarity change over training process}
        \label{fig:embedding change}
\end{figure}

Throughout the DPO process, our goal is to increase the difference between the ratios of conditional probabilities $\frac{\pi_{\theta}(y^+|x)}{\pi_{\text{ref}}(y^+|x)}$ and $\frac{\pi_{\theta}(y^-|x)}{\pi_{\text{ref}}(y^-|x)}$. The parameterized conditional probability $\pi_{\theta}(y|x)$ is changing throughout training. Thus, the embedding of a joint string $(x,y)$ changes as well. As shown in the upper part of Figure~\ref{fig:embed diff}, the components of the joint string embedding $(x,y)$ obviously change during DPO finetuning.  The characteristics of the response pairs will change over the course of DPO alignment, which implies that the selection and hard/easy labeling have to be done many times throughout the training process.

Alternatively, we only focus on the embedding of the response string $y$. As shown in the lower part of Figure~\ref{fig:embed diff}, the response embedding changes slightly during the DPO process. This is likely because the internal representation has been formed throughout the pre-training stage with a large document corpus.

Thus, we further calculate the cosine similarity of some batch of response pairs covering DPO finetuning.  As shown in Figure~\ref{fig:embed pair},  the distribution of cosine similarity between two independent responses does not change throughout training. This empirical observation that response embeddings stay consistent throughout the DPO finetuning is crucial since we can select responses before finetuning in the off-policy setting.


\section{Experiments}

\subsection{Dataset}
For the experiments, we use \textbf{SHP2} \cite{ethayarajh22a} dataset $\{y^1_i,y^2_i,\ldots,y^K_i,x_i\}_{i=1}^N$, which contains 4.8M prompts and response pairs. Both prompts and responses are collected from social media posts such as Reddit. Each response has a score, which is calculated from the number of positive and negative votes. To show that our results are not dataset-specific, we also consider the standard LLM alignment datasets within \textbf{Anthropic HH-RLHF}: helpfulness base and harmlessness base \cite{bai2022training}. They follow the classic data setting of $\{y^+_i,y^-_i,x_i\}_{i=1}^N$ and consist of 43k and 44k prompts and corresponding response pairs, generated primarily by proprietary language models trained by Anthropic and not released to the public. Appendix \ref{sec:A} describes more data details. 

\subsection{Experiment setting}
\begin{figure*}[h]
     \centering
          \begin{subfigure}[h]{0.64\textwidth}
         \centering
         \includegraphics[width=\textwidth]{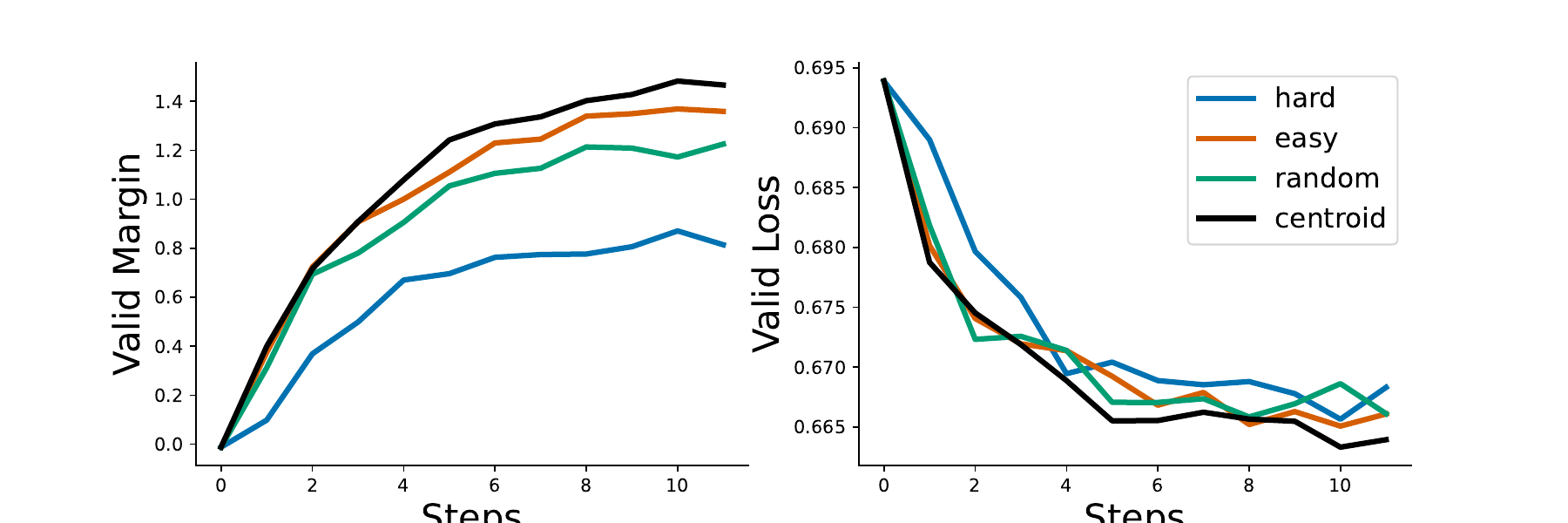}
         \caption{Margins and Loss on the SHP2 test dataset}
         \label{fig:shp_margins}
     \end{subfigure}
          \begin{subfigure}[h]{0.35\textwidth}
         \centering
         \includegraphics[width=\textwidth]{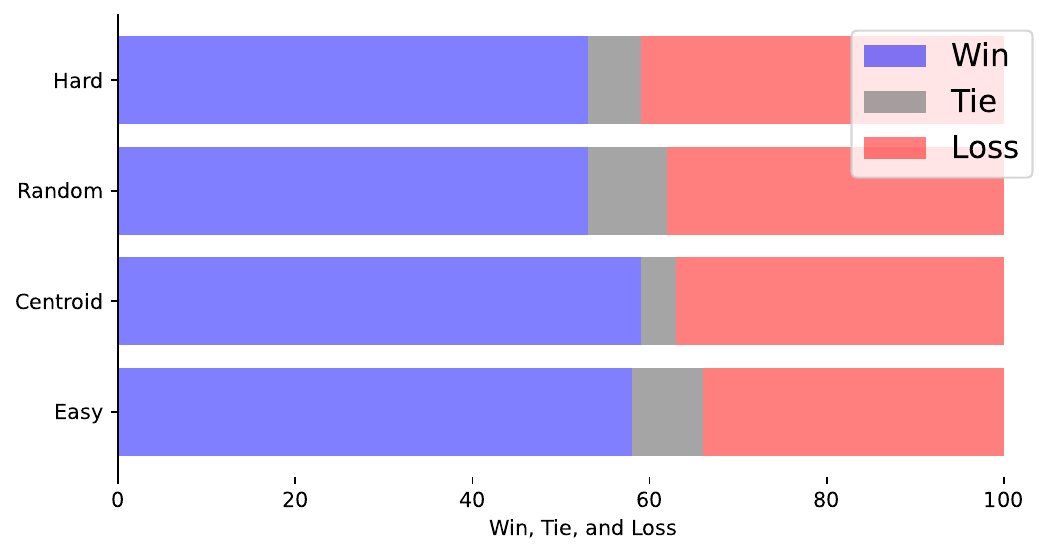}
         \caption{Evaluate the response using GPT4 by comparing to the reference model}
         \label{winrate_shp}
     \end{subfigure}
        \caption{(a) Margins and Loss on the SHP2 Test Dataset, b) Evaluation of Responses Using GPT-4 Compared to the reference model.}
        \label{fig:shps}
\end{figure*}
Three different sizes of base LLMs (Phi1.5~\cite{textbooks2}, Pythia2.8B~\cite{biderman2023pythia}, Llama2 7B~\cite {touvron2023llama}) are aligned with DPO using the differently selected subdata. All experiments were run on 4 NVIDIA A100 GPUs for one epoch to prevent overfitting. For more technical details, see appendix \ref{app c}.
To evaluate the checkpoints on the test dataset $\mathcal{D}_{\text{test}}$ over the alignment process. Following \cite{yu2024direct,xia2024less}, we employ loss and margins as the target metrics and use the implicit reward score extracted from the model to calculate them. Given a prompt $x$ and the corresponding chosen and rejected responses $y^+$ and $y^-,$ the margins are defined as
\begin{equation}\small
    \text{Margins} = \frac{1}{|\mathcal{D}_{\text{test}}|}\sum_{x,y^+,y^-} r(x,y^+)- r(x,y^-)
\end{equation}
and the DPO loss is defined as
\begin{equation}\small
    \text{Loss} = - \frac{1}{|\mathcal{D}_{\text{test}}|}\sum_{x,y^+,y^-} \log \sigma(\beta r(x,y^+)- \beta r(x,y^-))),
\end{equation}
where the implicit reward $r(x, y)$ is calculated from the reference and aligned models probability outputs as $r(x,y) = \frac{\pi_{\theta}(y|x)}{\pi_{\text{ref}}(y|x)}$. We use both margins and loss to analyze the DPO results with a high degree of confidence.


\subsection{Experiment on SHP2}
We used \textbf{SHP2} Reddit folder, where each prompt is a Reddit post with a question/instruction, and the responses are the top-level comments for that post. After clearing, each prompt has three responses, i.e., $\{y^1_i,y^2_i,y^3_i,x_i\}_{i=1}^N$. We use Eq.~\ref{eq:comp_hard}, Eq.~\ref{eq:comp_easy}, Eq.~\ref{eq:centroid} to construct the ``\textbf{hard}'', ``\textbf{easy}'', and ``\textbf{centroid}'' sub-dataset, respectively. When $K=3$, ``\textbf{centroid}'' sub-dataset is the mix dataset of ``\textbf{easy}'' and ``\textbf{random}''. The details of the experiment setting can be found in Appendix~\ref{app c}.




\begin{table}[h!]\small
\centering
  \caption{Learning Margins on SHP2 dataset.}
\begin{tabularx}{0.99\columnwidth}{ccccm{1.5cm}}
\Xhline{2\arrayrulewidth}
& {\textbf{Centroid}}& {\textbf{Easy}}&{\textbf{Random}}&{\textbf{Hard}}\\
 \hline
Phi1.5 &\textbf{1.41} &1.22&1.18&0.78\\
Pythia &\textbf{2.54} &2.33&2.23&1.88\\
Llama2 &\textbf{3.65} &3.28&3.18&2.85\\
\Xhline{2\arrayrulewidth}
\end{tabularx}
\label{tab:shp2_result}
\end{table}

\subsubsection{Results}
First, we finetune the base model based on the training split of the SHP2 dataset. Due to the large dataset size, we randomly select $1/3$ data points (with different random seeds) from the whole training data to do the SFT. Using the SFT model as the reference, we conduct the DPO alignment with four selected subsets of response pairs.

 As shown in Fig.~\ref{fig:shp_margins}, the ``\textbf{hard}'' has smaller margins and larger losses compared to others. Both Easy and Centroid checkpoints show good results compared to the baseline Random.  By mixing ``\textbf{easy}'' and ``\textbf{random}'', ``\textbf{centroid}'' obtained a better result on margins than Easy, as shown in Table \ref{tab:shp2_result}. 

To evaluate the aligned models, we randomly select $100$ prompts from the SHP2 test split and generate the responses using five versions of the checkpoints: SFT, ``\textbf{random}'', ``\textbf{hard}'', ``\textbf{easy}'', and ``\textbf{centroid}''. We use AlpacaEval-GPT4 to judge the relative alignment of the test responses to human preferences. Specifically, we directly compare the responses of such checkpoints to the responses of the SFT reference model to calculate the win rate depicted in Fig.~\ref{winrate_shp}.
''\textbf{centroid}'' has the best Win rate, and Easy has the biggest Loss rate. This experiment provides evidence that using centroid and ``\textbf{easy}'' data to train the model by DPO will lead to a more safety and helpful generated model. Table \ref{tab:generative result} (in Appendix~\ref{app:D}) shows the generative responses.

\begin{figure*}[h!]
     \centering
     \begin{subfigure}[h!]{0.85\textwidth}
         \centering
         \includegraphics[width=\textwidth]{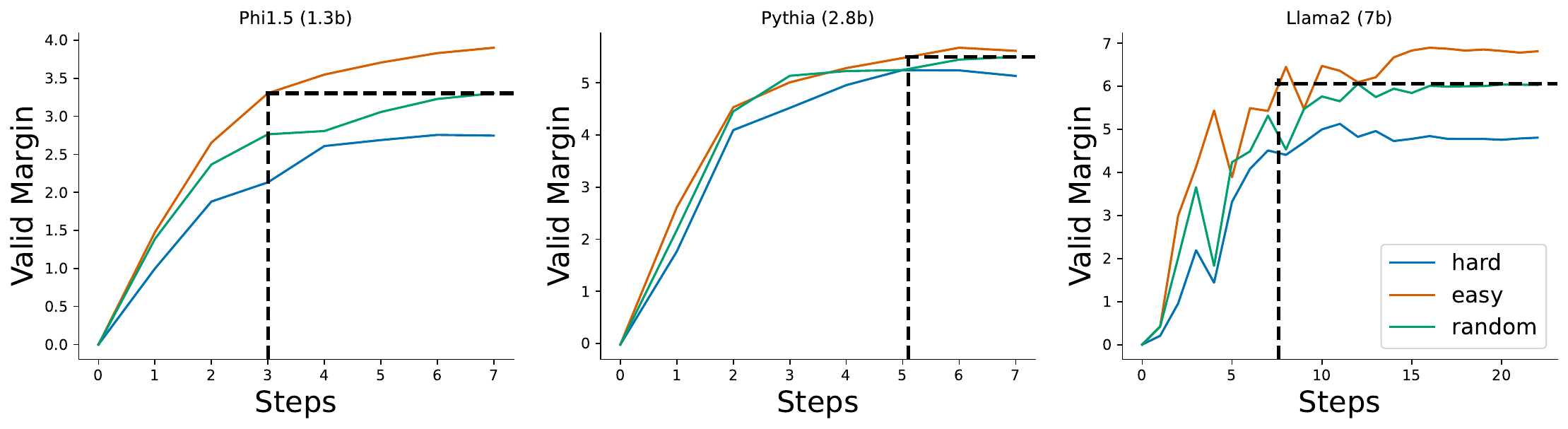}
         \caption{Evaluate margins}
         \label{fig:eval-margin}
     \end{subfigure}
     \begin{subfigure}[h!]{0.85\textwidth}
         \centering
         \includegraphics[width=\textwidth]{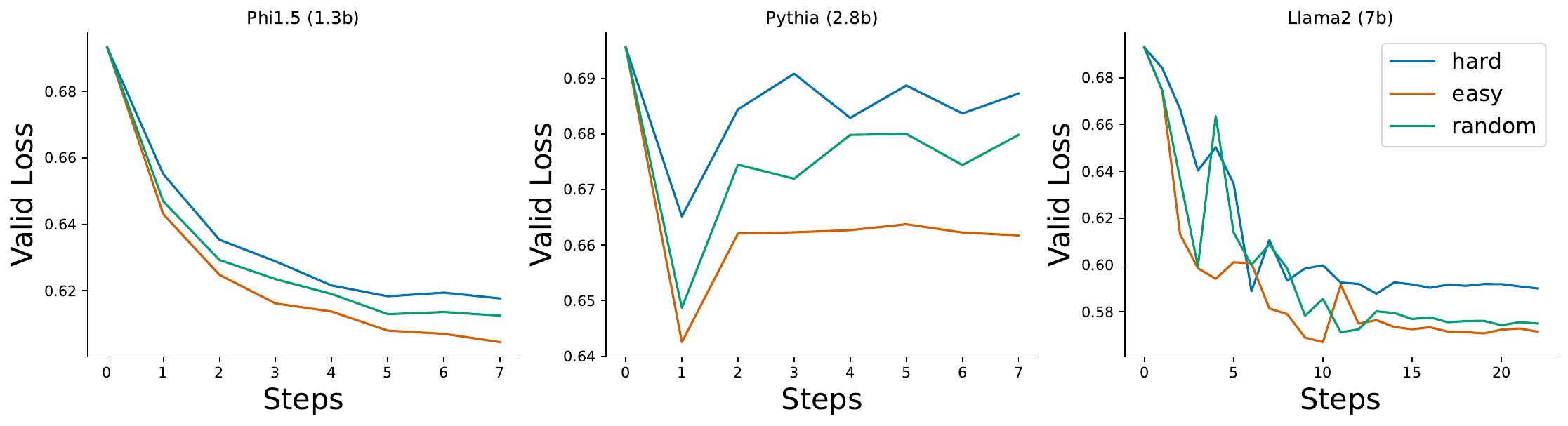}
         \caption{Evaluate loss}
         \label{fig:eva-loss}
     \end{subfigure}
        \caption{Result of evaluating margins and loss on different models}
        \label{fig:eva-test}
\end{figure*}

\begin{table*}[h!]
\centering
  \caption{Win Rate of the aligned model when compared against the reference supervised finetuned model. $90\%$ margin of error is $0.0074.$ }
\begin{tabular}{ccccccc|cccccc}
\Xhline{2\arrayrulewidth}
 \multirow{3}{*}{\textbf{Sub-set}}&\multicolumn{6}{c|}{\textbf{Harmlessness}}& \multicolumn{6}{c}{\textbf{Helpfulness}}\\
\setlength{\arrayrulewidth}{0.2pt}
&\multicolumn{2}{c}{\textbf{Phi-1.3B}}&\multicolumn{2}{c}{\textbf{Pythia-2.8B}}& \multicolumn{2}{c}{\textbf{Llama2-7B}}&\multicolumn{2}{|c}{\textbf{Phi-1.3B}}&\multicolumn{2}{c}{\textbf{Pythia-2.8B}}& \multicolumn{2}{c}{\textbf{Llama2-7B}}\\
  & \textbf{Win} $\uparrow$ & \textbf{Lose}$\downarrow$ & \textbf{Win}$\uparrow$ &\textbf{Lose}$\downarrow$& \textbf{win}$\uparrow$ &\textbf{Lose}$\downarrow$& \textbf{Win}$\uparrow$ & \textbf{Lose}$\downarrow$ & \textbf{Win}$\uparrow$ &\textbf{Lose}$\downarrow$& \textbf{win}$\uparrow$ &\textbf{Lose}$\downarrow$ \\
 \hline
Random &0.54 &0.46& 0.64 &0.36&0.67 &0.33&0.71 &0.29& 0.71 &\textbf{0.26}&0.76 &0.24\\
Hard &0.53 &0.47 & 0.65&0.32&0.66&0.34&0.72 &0.28 & 0.66&0.34&0.78&0.21\\
Easy &\textbf{0.55} &\textbf{0.45}&\textbf{0.69} &\textbf{0.30} &\textbf{0.72}&\textbf{0.28}&\textbf{0.76} &\textbf{0.24}&\textbf{0.72} &{0.27} &\textbf{0.80}&\textbf{0.19}\\
\Xhline{2\arrayrulewidth}
\end{tabular}
\label{result_hh}
\end{table*}


\subsection{Experiment on Anthropic HH-RLHF}

Additionally, we explore the effect of a similar/dissimilar pair for the standard alignment dataset Anthropic HH-RLHF datasets consist of one response pair for each prompt, i.e, $\{y^1_i,y^2_i,x_i\}_{i=1}^N$, after removing the labels. 
We sort all data $\{y^1_i,y^2_i,x_i\}_{i=1}^N$ by descending based on the cosine similarity $\cos(\boldsymbol{y}^1_i,\boldsymbol{y}^2_i)$ to obtain the sorted dataset $\mathcal{D}_{\text{sort}} = \{y^1_s,y^2_s,x_s\}_{s=1}^N$
Thus, the $\mathcal{D}_{\text{sort}}$ is sorted from the hardest (most similar) pair to the easiest (least similar) pair. We define $\mathcal{D}_{\text{hard}}=\{y^1_s,y^2_s,x_s\}_{s=1}^{N/2}$ and $\mathcal{D}_{\text{easy}}=\{y^1_s,y^2_s,x_s\}_{s=N/2+1}^{N}$ as the first half and the second half of $\mathcal{D}_{\text{sort}}$ respectively.

We randomly select 50\% of the tuples from the entire Anthropic HH-RLHF as the ``\textbf{random}'' baseline dataset. The  ``\textbf{random}'' dataset represents the data distribution that would be the default for a state-of-the-art alignment system. The details of the experiment setting can be found in Appendix~\ref{app c}.

\subsubsection{Results}

We supervised finetuned (SFT) the base model using the chosen response from the training split of the entire dataset to get the $\pi_{\text{ref}}$. Then we use ``\textbf{hard}'' and  ``\textbf{easy}'' and ``\textbf{random}'' sub-sets to align the SFT model $\pi_{\text{ref}}$ using the DPO procedure. 

We depict the dynamics of the loss on the test dataset throughout training in Fig.~\ref{fig:eva-loss} and margin in Fig.~\ref{fig:eval-margin}. For both HH-harmlessness and HH-helpfulness datasets, training on the ``\textbf{easy}'' sub-set results in better values of the Loss and Margins than training on the ``\textbf{random}'' and ``\textbf{hard}'' sub-sets. On the contrary, alignment on the ``\textbf{hard}'' subset hurts the performance when compared to the baseline trained on the ``\textbf{random}'' subset.  The dashed line in Fig.~\ref{fig:eval-margin} demonstrates sample efficiency of the alignment on the easy subset when compared to the ``\textbf{random}'' subset baseline. The proposed approach of adaptive data selection saves from $28\%$ to $65\%$ of labeling data efforts. 
    
We also compare the final checkpoints with the Win Rate evaluation metric. We analyze win rates using GPT-4 to judge the responses of an aligned model against the responses of the SFT reference model before alignment. In this experiment, we used $100$ of the prompts sampled from the test subsets of the Harmlessness and Helpfulness datasets. In this experiment, there were $100$ sample pairs, and the standard error ranged from $0.042$ to $0.045$, resulting in a $90\%$ margin of error of $0.0074$.

 We use the AlpacaEval-GPT4\footnote{Licensed under the Apache License, Version 2.0} \cite{dubois2024length} framework to assess the judgment of the responses. The prompt supplied to the judge model is demonstrated in Appendix~\ref{app:D}. It was constructed to rank the responses according to human preferences into more preferred, less preferred, and a tie. The results of the comparison are shown in Table~\ref{result_hh}. The best results are provided in bold font. For both data evaluations, the response generated from the model aligned on the ``\textbf{easy}'' has a higher win rate and a lower loss rate than the other two models.  The ``\textbf{hard}'' response examples could hurt the performance, as evident from the Win Rate numbers as well. These results are consistent with the conclusions made from the loss and margin metrics.

\begin{figure}[h!]
     \centering
         \includegraphics[width=0.45\textwidth]{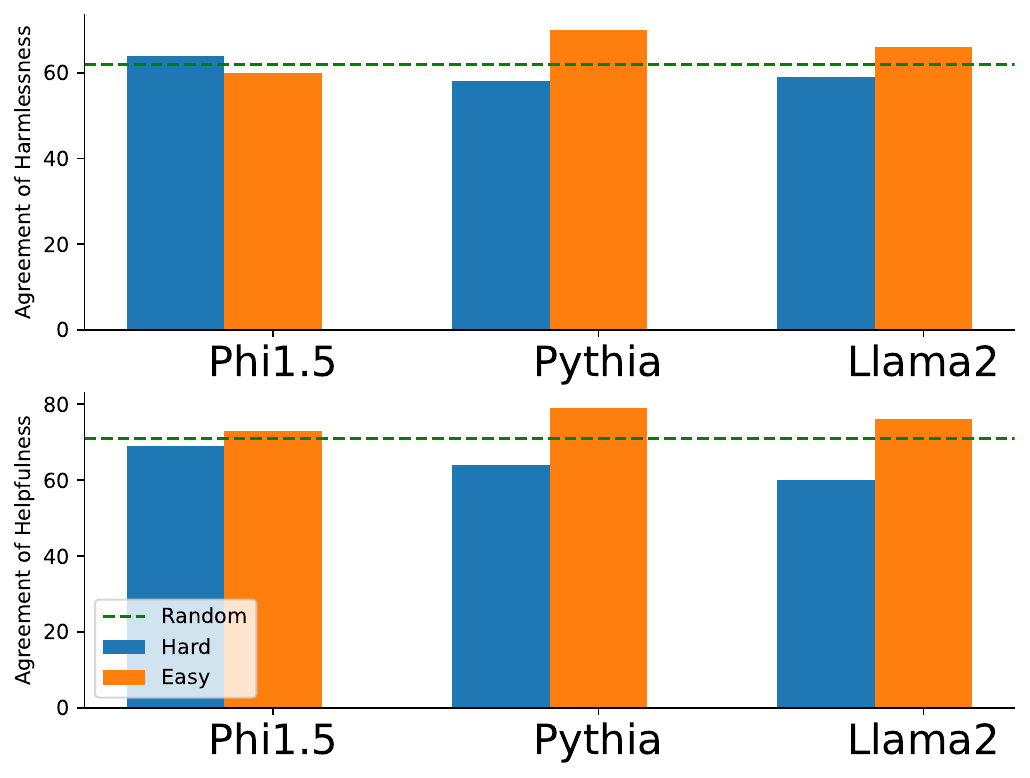}
        \caption{The percentage of response pairs labeled by GPT4 in the same way as they were initially labeled within Anthropic-HH.}
        \label{fig:wr_bias}
\end{figure}


\subsection{Discussion: Errors in the labeling}

We further interpret the superiority of the ``\textbf{easy}'' sub-set for alignment purposes.
The reason for the superior quality of the ``\textbf{easy}'' alignment sub-set might be the decreased frequency of erroneous and biased data points within the training dataset.
In the dataset, we assume that the chosen response $y^+$ will be more preferred by the human than the rejected response $y^-$ so $y^+ \succ y^-$. However, excessive reliance inevitably leads to errors and impedes learning if the initial annotations are inherently noisy ($y^- \succ y^+$). In contrastive learning, such data points would be called ``false negatives''~\cite{zhang2024hasa}. We randomly selected $200$ response pairs from ``\textbf{easy}'' and ``\textbf{hard}'' sub-sets sorted using the embeddings from all three base models. We formed a baseline by randomly choosing $200$ pairs from the Helpfulness and Harmlessness datasets. The GPT4 was used to re-label the pairs into chosen and rejected responses. In Figure \ref{fig:wr_bias}, we plot the proportion of response pairs that were assigned the same labels as the ones recorded in the original dataset. The ``\textbf{hard}'' subset, which consists of more similar response pairs, evidently also contains more response pairs where the ranking is unclear, resulting in more incorrect pairs than the ``\textbf{random}'' and ``\textbf{easy}'' subsets. Similarly, the ``\textbf{easy}'' subset contains more confidently ranked pairs and fewer erroneous response labels, which results in a higher dataset quality. $\mathcal{D}_{\text{easy}}$ has more confidence that sample from optimal preference distribution $p^{*}$ than $\mathcal{D}_{\text{hard}}$. It is consistent with the research that filters similar samples such as Llama3 \cite{dubey2024llama} and the observation that hard will bring more false negatives in contrastive learning~\cite{zhang2024hasa}. For LLM alignment, the effect of false hard examples is so drastic that it negates the positive effect of hard examples.

\section{Conclusion} 
This paper investigates a data selection method for DPO alignment in the embedding space. By considering the cosine similarity of response pairs independent of prompt variations, we propose a high-quality and least ambiguous data selection strategy. Our findings suggest that dissimilar response pairs more effectively align LLMs with human preferences. In contrast, highly similar response pairs are more prone to label noise, which can negatively impact DPO alignment. We evaluated our approach in both the real-world dataset SHP2 and the synthetic dataset HH-RLHF, demonstrating its effectiveness. It can extend to online-DPO. 

\section*{Limitation}
In this paper, we only finetune the 1.3B to 7B language model. The bigger language model, 70B, is hard to implement due to resource limitations. Thus, for model sizes over 10B, our method needs to be further discussed while the Llama3 model filters the similar pairs in their report~\cite{dubey2024llama}.

For the metric of Win Rate, we use the GPT4 as the annotator without considering the inherent bias compared to the Human annotator. We report the margin of error to provide confidence of the report values. 

\section*{Acknowledgements}
This work was supported by the National Science Foundation AI Institute in Dynamic Systems (Grant No. 2112085) and NRT-AI 2244574.
This work used cloud GPU resources at NCSA Delta cluster through allocation number CIS240027 from the Advanced Cyberinfrastructure Coordination Ecosystem: Services \& Support (ACCESS) program, which is supported by National Science Foundation grants \#2138259, \#2138286, \#2138307, \#2137603, and \#2138296.
This research was also generously supported in part by the Google Cloud Research Credits Program with the award GCP19980904 and OpenAI Researcher Access Program.

\bibliographystyle{named}
\bibliography{ijcai25}

\begin{thebibliography}{}

\bibitem[\protect\citeauthoryear{Azar \bgroup \em et al.\egroup }{2023}]{azar2023general}
Mohammad~Gheshlaghi Azar, Mark Rowland, Bilal Piot, Daniel Guo, Daniele Calandriello, Michal Valko, and R{\'e}mi Munos.
\newblock A general theoretical paradigm to understand learning from human preferences.
\newblock {\em arXiv preprint arXiv:2310.12036}, 2023.

\bibitem[\protect\citeauthoryear{Bai \bgroup \em et al.\egroup }{2022}]{bai2022training}
Yuntao Bai, Andy Jones, Kamal Ndousse, Amanda Askell, Anna Chen, Nova DasSarma, Dawn Drain, Stanislav Fort, Deep Ganguli, Tom Henighan, et~al.
\newblock Training a helpful and harmless assistant with reinforcement learning from human feedback.
\newblock {\em arXiv preprint arXiv:2204.05862}, 2022.

\bibitem[\protect\citeauthoryear{Biderman \bgroup \em et al.\egroup }{2023}]{biderman2023pythia}
Stella Biderman, Hailey Schoelkopf, Quentin~Gregory Anthony, Herbie Bradley, Kyle O’Brien, Eric Hallahan, Mohammad~Aflah Khan, Shivanshu Purohit, USVSN~Sai Prashanth, Edward Raff, et~al.
\newblock Pythia: A suite for analyzing large language models across training and scaling.
\newblock In {\em International Conference on Machine Learning}, pages 2397--2430. PMLR, 2023.

\bibitem[\protect\citeauthoryear{Boerner \bgroup \em et al.\egroup }{2023}]{Boerner2023ACCESS}
Timothy~J. Boerner, Stephen Deems, Thomas~R. Furlani, Shelley~L. Knuth, and John Towns.
\newblock {ACCESS: Advancing Innovation: NSF’s Advanced Cyberinfrastructure Coordination Ecosystem: Services \& Support}.
\newblock In {\em Proceedings of the Practice and Experience in Advanced Research Computing (PEARC '23)}, page~4, New York, NY, USA, July 23--27 2023. ACM.

\bibitem[\protect\citeauthoryear{Bradley and Terry}{1952}]{bradley1952rank}
Ralph~Allan Bradley and Milton~E Terry.
\newblock Rank analysis of incomplete block designs: I. the method of paired comparisons.
\newblock {\em Biometrika}, 39(3/4):324--345, 1952.

\bibitem[\protect\citeauthoryear{Chen \bgroup \em et al.\egroup }{2020}]{chen2020simple}
Ting Chen, Simon Kornblith, Mohammad Norouzi, and Geoffrey Hinton.
\newblock A simple framework for contrastive learning of visual representations.
\newblock In {\em International conference on machine learning}, pages 1597--1607. PMLR, 2020.

\bibitem[\protect\citeauthoryear{Christiano \bgroup \em et al.\egroup }{2017}]{christiano2017deep}
Paul~F Christiano, Jan Leike, Tom Brown, Miljan Martic, Shane Legg, and Dario Amodei.
\newblock Deep reinforcement learning from human preferences.
\newblock {\em Advances in neural information processing systems}, 30, 2017.

\bibitem[\protect\citeauthoryear{Chuang \bgroup \em et al.\egroup }{2020}]{chuang2020debiased}
Ching-Yao Chuang, Joshua Robinson, Yen-Chen Lin, Antonio Torralba, and Stefanie Jegelka.
\newblock Debiased contrastive learning.
\newblock {\em Advances in neural information processing systems}, 33:8765--8775, 2020.

\bibitem[\protect\citeauthoryear{Devlin}{2018}]{devlin2018bert}
Jacob Devlin.
\newblock Bert: Pre-training of deep bidirectional transformers for language understanding.
\newblock {\em arXiv preprint arXiv:1810.04805}, 2018.

\bibitem[\protect\citeauthoryear{Dubey \bgroup \em et al.\egroup }{2024}]{dubey2024llama}
Abhimanyu Dubey, Abhinav Jauhri, Abhinav Pandey, Abhishek Kadian, Ahmad Al-Dahle, Aiesha Letman, Akhil Mathur, Alan Schelten, Amy Yang, Angela Fan, et~al.
\newblock The llama 3 herd of models.
\newblock {\em arXiv preprint arXiv:2407.21783}, 2024.

\bibitem[\protect\citeauthoryear{Dubois \bgroup \em et al.\egroup }{2024}]{dubois2024length}
Yann Dubois, Bal{\'a}zs Galambosi, Percy Liang, and Tatsunori~B Hashimoto.
\newblock Length-controlled alpacaeval: A simple way to debias automatic evaluators.
\newblock {\em arXiv preprint arXiv:2404.04475}, 2024.

\bibitem[\protect\citeauthoryear{Ethayarajh \bgroup \em et al.\egroup }{2022}]{ethayarajh22a}
Kawin Ethayarajh, Yejin Choi, and Swabha Swayamdipta.
\newblock Understanding dataset difficulty with $\mathcal{V}$-usable information.
\newblock In Kamalika Chaudhuri, Stefanie Jegelka, Le~Song, Csaba Szepesvari, Gang Niu, and Sivan Sabato, editors, {\em Proceedings of the 39th International Conference on Machine Learning}, volume 162 of {\em Proceedings of Machine Learning Research}, pages 5988--6008. PMLR, 17--23 Jul 2022.

\bibitem[\protect\citeauthoryear{Guo \bgroup \em et al.\egroup }{2024}]{guo2024direct}
Shangmin Guo, Biao Zhang, Tianlin Liu, Tianqi Liu, Misha Khalman, Felipe Llinares, Alexandre Rame, Thomas Mesnard, Yao Zhao, Bilal Piot, et~al.
\newblock Direct language model alignment from online ai feedback.
\newblock {\em arXiv preprint arXiv:2402.04792}, 2024.

\bibitem[\protect\citeauthoryear{Jiang \bgroup \em et al.\egroup }{2023}]{jiang2023scaling}
Ting Jiang, Shaohan Huang, Zhongzhi Luan, Deqing Wang, and Fuzhen Zhuang.
\newblock Scaling sentence embeddings with large language models.
\newblock {\em arXiv preprint arXiv:2307.16645}, 2023.

\bibitem[\protect\citeauthoryear{Li \bgroup \em et al.\egroup }{2023}]{textbooks2}
Yuanzhi Li, S{\'e}bastien Bubeck, Ronen Eldan, Allie Del~Giorno, Suriya Gunasekar, and Yin~Tat Lee.
\newblock Textbooks are all you need ii: \textbf{phi-1.5} technical report.
\newblock {\em arXiv preprint arXiv:2309.05463}, 2023.

\bibitem[\protect\citeauthoryear{Liu \bgroup \em et al.\egroup }{2023}]{liu2023chain}
Hao Liu, Carmelo Sferrazza, and Pieter Abbeel.
\newblock Chain of hindsight aligns language models with feedback.
\newblock {\em arXiv preprint arXiv:2302.02676}, 2023.

\bibitem[\protect\citeauthoryear{Mikolov}{2013}]{mikolov2013efficient}
Tomas Mikolov.
\newblock Efficient estimation of word representations in vector space.
\newblock {\em arXiv preprint arXiv:1301.3781}, 2013.

\bibitem[\protect\citeauthoryear{Morimura \bgroup \em et al.\egroup }{2024}]{morimura2024filtered}
Tetsuro Morimura, Mitsuki Sakamoto, Yuu Jinnai, Kenshi Abe, and Kaito Air.
\newblock Filtered direct preference optimization.
\newblock {\em arXiv preprint arXiv:2404.13846}, 2024.

\bibitem[\protect\citeauthoryear{Ouyang \bgroup \em et al.\egroup }{2022}]{ouyang2022training}
Long Ouyang, Jeffrey Wu, Xu~Jiang, Diogo Almeida, Carroll Wainwright, Pamela Mishkin, Chong Zhang, Sandhini Agarwal, Katarina Slama, Alex Ray, et~al.
\newblock Training language models to follow instructions with human feedback.
\newblock {\em Advances in neural information processing systems}, 35:27730--27744, 2022.

\bibitem[\protect\citeauthoryear{Rafailov \bgroup \em et al.\egroup }{2024}]{rafailov2024direct}
Rafael Rafailov, Archit Sharma, Eric Mitchell, Christopher~D Manning, Stefano Ermon, and Chelsea Finn.
\newblock Direct preference optimization: Your language model is secretly a reward model.
\newblock {\em Advances in Neural Information Processing Systems}, 36, 2024.

\bibitem[\protect\citeauthoryear{Schulman \bgroup \em et al.\egroup }{2017}]{schulman2017proximal}
John Schulman, Filip Wolski, Prafulla Dhariwal, Alec Radford, and Oleg Klimov.
\newblock Proximal policy optimization algorithms.
\newblock {\em arXiv preprint arXiv:1707.06347}, 2017.

\bibitem[\protect\citeauthoryear{Settles}{2009}]{settles2009active}
Burr Settles.
\newblock Active learning literature survey.
\newblock 2009.

\bibitem[\protect\citeauthoryear{Song \bgroup \em et al.\egroup }{2023}]{song2023preference}
Feifan Song, Bowen Yu, Minghao Li, Haiyang Yu, Fei Huang, Yongbin Li, and Houfeng Wang.
\newblock Preference ranking optimization for human alignment.
\newblock {\em arXiv preprint arXiv:2306.17492}, 2023.

\bibitem[\protect\citeauthoryear{Stiennon \bgroup \em et al.\egroup }{2020}]{stiennon2020learning}
Nisan Stiennon, Long Ouyang, Jeffrey Wu, Daniel Ziegler, Ryan Lowe, Chelsea Voss, Alec Radford, Dario Amodei, and Paul~F Christiano.
\newblock Learning to summarize with human feedback.
\newblock {\em Advances in Neural Information Processing Systems}, 33:3008--3021, 2020.

\bibitem[\protect\citeauthoryear{Touvron \bgroup \em et al.\egroup }{2023}]{touvron2023llama}
Hugo Touvron, Louis Martin, Kevin Stone, Peter Albert, Amjad Almahairi, Yasmine Babaei, Nikolay Bashlykov, Soumya Batra, Prajjwal Bhargava, Shruti Bhosale, et~al.
\newblock Llama 2: Open foundation and fine-tuned chat models.
\newblock {\em arXiv preprint arXiv:2307.09288}, 2023.

\bibitem[\protect\citeauthoryear{Wang \bgroup \em et al.\egroup }{2024a}]{wang2024secrets}
Binghai Wang, Rui Zheng, Lu~Chen, Yan Liu, Shihan Dou, Caishuang Huang, Wei Shen, Senjie Jin, Enyu Zhou, Chenyu Shi, et~al.
\newblock Secrets of rlhf in large language models part ii: Reward modeling.
\newblock {\em arXiv preprint arXiv:2401.06080}, 2024.

\bibitem[\protect\citeauthoryear{Wang \bgroup \em et al.\egroup }{2024b}]{wang2024essence}
Xinpeng Wang, Shitong Duan, Xiaoyuan Yi, Jing Yao, Shanlin Zhou, Zhihua Wei, Peng Zhang, Dongkuan Xu, Maosong Sun, and Xing Xie.
\newblock On the essence and prospect: An investigation of alignment approaches for big models.
\newblock {\em arXiv preprint arXiv:2403.04204}, 2024.

\bibitem[\protect\citeauthoryear{Wu \bgroup \em et al.\egroup }{2024}]{wu2024beta}
Junkang Wu, Yuexiang Xie, Zhengyi Yang, Jiancan Wu, Jinyang Gao, Bolin Ding, Xiang Wang, and Xiangnan He.
\newblock $\beta$-dpo: Direct preference optimization with dynamic $\beta$.
\newblock {\em arXiv preprint arXiv:2407.08639}, 2024.

\bibitem[\protect\citeauthoryear{Xia \bgroup \em et al.\egroup }{2024}]{xia2024less}
Mengzhou Xia, Sadhika Malladi, Suchin Gururangan, Sanjeev Arora, and Danqi Chen.
\newblock Less: Selecting influential data for targeted instruction tuning.
\newblock {\em arXiv preprint arXiv:2402.04333}, 2024.

\bibitem[\protect\citeauthoryear{Xu \bgroup \em et al.\egroup }{2023}]{xu2023contrastive}
Canwen Xu, Corby Rosset, Luciano Del~Corro, Shweti Mahajan, Julian McAuley, Jennifer Neville, Ahmed~Hassan Awadallah, and Nikhil Rao.
\newblock Contrastive post-training large language models on data curriculum.
\newblock {\em arXiv preprint arXiv:2310.02263}, 2023.

\bibitem[\protect\citeauthoryear{Yu \bgroup \em et al.\egroup }{2024}]{yu2024direct}
Runsheng Yu, Yong Wang, Xiaoqi Jiao, Youzhi Zhang, and James~T Kwok.
\newblock Direct alignment of language models via quality-aware self-refinement.
\newblock {\em arXiv preprint arXiv:2405.21040}, 2024.

\bibitem[\protect\citeauthoryear{Zeng \bgroup \em et al.\egroup }{2024}]{zeng2024token}
Yongcheng Zeng, Guoqing Liu, Weiyu Ma, Ning Yang, Haifeng Zhang, and Jun Wang.
\newblock Token-level direct preference optimization.
\newblock {\em arXiv preprint arXiv:2404.11999}, 2024.

\bibitem[\protect\citeauthoryear{Zhang \bgroup \em et al.\egroup }{2024}]{zhang2024hasa}
Honggen Zhang, June Zhang, and Igor Molybog.
\newblock Hasa: Hardness and structure-aware contrastive knowledge graph embedding.
\newblock In {\em Proceedings of the ACM on Web Conference 2024}, pages 2116--2127, 2024.

\bibitem[\protect\citeauthoryear{Zhao \bgroup \em et al.\egroup }{2022}]{zhao2022calibrating}
Yao Zhao, Misha Khalman, Rishabh Joshi, Shashi Narayan, Mohammad Saleh, and Peter~J Liu.
\newblock Calibrating sequence likelihood improves conditional language generation.
\newblock {\em arXiv preprint arXiv:2210.00045}, 2022.

\bibitem[\protect\citeauthoryear{Zhao \bgroup \em et al.\egroup }{2023}]{zhao2023slic}
Yao Zhao, Rishabh Joshi, Tianqi Liu, Misha Khalman, Mohammad Saleh, and Peter~J Liu.
\newblock Slic-hf: Sequence likelihood calibration with human feedback.
\newblock {\em arXiv preprint arXiv:2305.10425}, 2023.

\end{thebibliography}
\appendix
\section{Appendix A: Related work}

\textbf{Direct Alignment of Language Models:} Despite RLHF’s effectiveness in aligning
language models (LMs) with human values, its complexity and resource demands have spurred the exploration of alternatives. Sequence Likelihood Calibration (SLiC)\cite{zhao2022calibrating} is a DAP method to directly encourage the LLMs to output positive responses and penalize negative responses. Chain of Hindsight (CoH) \cite{liu2023chain} is equivalent to learning a conditional policy.
DPO \cite{rafailov2024direct} directly optimizes LMs using a preference-based loss function to enhance training stability in comparison to traditional RLHF. DPO with Dynamic $\beta$ \cite{wu2024beta} introduced a framework that dynamically calibrates $\beta$ at the batch level, informed by the underlying preference
data. Existing work~\cite{azar2023general} identified that DPO is susceptible to overfitting and introduced Identity Preference Optimization (IPO) as a solution to this issue. \cite{zeng2024token} noticed that the generative diversity of LLM deteriorated, and the KL divergence grew faster for less preferred responses compared with preferred responses, and they proposed token-level DPO (TDPO) to enhance the regulation of KL divergence.

\textbf{Data Quality for Direct Alignment}
DPO works with the off-policy setting and a static dataset. It needs a more curated and diverse static dataset to overcome the drawback of being easily overfitted in the specific domain. 
PRO \cite{song2023preference} proposed a preference ranking with listwise preference datasets with an augmented dataset.  However, finetuning was constrained by the limitations of available data. 
Contrastive Post-training \cite{xu2023contrastive} tries to build more datasets using other LLMs to advance the training process of DPO without considering the mistakes. Recently, similar to active learning to select samples based on current models \cite{settles2009active}, there exist works using AI as an annotator to monitor the quantity of data pairs for each training step, despite the cost being still expensive and the labeling quality~\cite{guo2024direct,morimura2024filtered}. \cite{yu2024direct} use LLMs to design a refinement function, which estimates the quality of positive and negative responses.
LESS \cite{xia2024less} is an optimizer-aware and practically efficient algorithm to estimate data influences on gradient similarity and search for instruction data selection.  However, this data selection is online, so it needs more computation. Recently, Llama3 \cite{dubey2024llama} has been boosted by filtering similar samples in the post-training stage, while it didn't report the technical details and rigorous analysis. However, none of the above methods considers the inherent mistake when annotating or labeling the preference pair when they augment the dataset. 

\section{Appendix B: Data Details}
\label{sec:A}
In this section, we list the dataset of HH-RLHF\footnote{\url{https://huggingface.co/datasets/Anthropic/hh-rlhf}} and SHP2\footnote{\url{https://huggingface.co/datasets/stanfordnlp/SHP-2}} we used for training and testing. 

HH Training dataset build step:
\begin{enumerate}
    \item Regardless of the prompt, we calculate the embedding similarity for all response pairs given the base model (Phi1.5, Pythia 2.8B LLaMa2-7B). Note that the embeddings are obtained by the average of tokens. 
    \item We rank the response pairs based on the similarity on the descent. Selecting the first $50\%$ response pairs as the Easy dataset and the rest of $50\%$ response pairs as the Easy dataset.
    \item We also random select the $50\%$ data as the Random dataset
\end{enumerate}
Thus, we will obtain three subdat. The Random and Easy, Hard will have $50\%$ overlapped. 

For the SHP2, we use the Reddit folder dataset. It is the response from the people's answer to the question from Reddit. Each response has a score which is evaluated by the positive votes and negative votes. 
We exclude the response pairs length over 512 and the ratio of chosen score and rejected score less than 1.5.  After that, we obtained 35.6k response pairs. Each prompt has three (most of) or four responses.  Training dataset build step:
\begin{enumerate}
    \item For each prompt, we calculate the embedding similarity of the combination of four or three responses. For example, the three responses will have $C^3_2 = 3$ response pairs.  
    \item We rank the $3$ response pairs based on the similarity of the descent. Selecting the largest similar value of response pairs as the Hard one and the 
    smallest similar value of response pairs as the Easy one. 
    \item We cluster the $3$ response based on two groups in the embedding space. We select the nearest centroid response as the center of the group. The centers will be regarded as the Centroid pairs. 
    \item We also random select two responses from the $3$ responses to build the Random dataset
\end{enumerate}
Thus, we will obtain four sub-datasets: Easy, Hard, Centroid, and Random. Note that Centroid has overlap with Random and Easy but Hard when the $k=3$.


\begin{table}[h!]
\centering
  \caption{Dataset description.}
\begin{tabularx}{0.9\columnwidth}{cccc} 
\hline
& {\textbf{Harmless}}& {\textbf{Helpful}}&{\textbf{SHP2}}\\
 \hline
Raw Data &42k &44k&35k\\
Train &21k &{22k}&11k\\
Test & 1.5k&1.5k&1.9k\\
\hline
\end{tabularx}
\label{tab:dataset}
\end{table}

\section{Appendix C}\label{appB}

\subsection{LLM Alignment to Human Feedback}

The human feedback rarely comes in the form of the true reward samples $(x, y, R_{xy}).$ Ranking or pairwise comparison of responses is more common. In the pairwise comparison case, there are two sample responses $(y^+,y^-)$ that correspond to a single prompt $x$. Here $y^+$ and $y^-$ are the preferred and non-preferred responses, respectively, as defined by
 human subjects providing a binary preference to label them $(R_{xy^+} > R_{xy^-} |x).$ This method of labeling does not explicitly provide access to the true reward function $R_{xy}$.
 However, alignment can still be performed by applying the Reinforcement Learning from Human Feedback (RLHF) algorithm using binary human preference data. 
 Human preference probability is conditioned on the prompt and the two versions of responses and could be defined as a mathematical expectation of the indicator that the first version of the response is going to be preferred over the second version by a random member of the human population.
 While the human preference distribution $p^{*}$ cannot be accessed, we always use the  
 the Bradley-Terry\cite{bradley1952rank} model to defines the log-odds associated with parameterized reward model $r(\cdot)$ 
\begin{equation}\label{eq: bt}
    \log \frac{p}{1-p} = r(x,y^+) - r(x,y^-),
\end{equation}
By optimizing the reward model, we will approximate the optimized human preference distribution $p^{*}$.
 Modeling the margins of the response pair can also be viewed from the perspective of estimating the true reward in Eq. \ref{eq:reward}. Assume the ground truth reward for $R_{xy^+}$ and $R_{xy^-}$ is 1 and 0 respectively. The difference between the estimated reward and the truth is $\mathbb{E}_{q(y^+)}[r(x,y^+) -1]+\mathbb{E}_{q(y^-)}[r(x,y^-) -0]  = \mathbb{E}[r(x,y^+)-r(x,y^-) -1]$. where $q(y^+)$ and $q(y^-)$ are the distribution of preference response and non-preference response, respectively. 

Arrangement the Eq.\ref{eq: bt}, we have
\small{
\begin{equation}
    p(y^+ \succ y^- |x) = \frac{\hbox{exp}(r(x,y^+))}{\hbox{exp}(r(x,y^+))+\hbox{exp}(r(x,y^-))}.
\end{equation}
}
In practice, we will learn the parametrized reward model given the human-labeled preference data. 
The reward model learned from the human preference responses can be used to score LLM-generated content.  It provides human feedback to the language model $\pi_{\theta}$ by maximizing the objective function 
\begin{align}
    \max_{\pi_{\theta}} & \mathbb{E}_{x\sim \mathcal{D},y\sim \pi_{\theta}(y|x)}[r(x,y)] \\ \nonumber & - \beta \mathbb{D}_{KL}[\pi_{\theta}(y|x)||\pi_{\text{ref}}(y|x)]
\end{align}
where $\pi_{\text{ref}}$ is the reference model after the supervised fine-tuning.


\section{Appendix D: Experiments Setting:}\label{app c}
 We use three different-sized base LLMs (Phi1.5-1.3B, Pythia 2.8B, and Llama2 7B) as the backbone in our experiments. Using the dataset of the preferred prompt-response pairs $(x, y^+)$, we run the supervised finetuning (SFT) procedure with each base model to obtain their SFT versions. The supervised finetuning does not require a pairwise comparison of the responses and can be performed on prompt-response pairs outside the alignment dataset. However, our experiments use the preferred response pairs from the dataset under consideration (Anthropic HH-RLHF or SHP2). After the SFT stage, we run separate DPO procedures on each of the selected sub-sets (easy, hard, ect.) with each of the three SFT LLMs. This procedure is illustrated in Algorithm \ref{alg:align}. All experiments are run on 4 A100 GPUs \cite{Boerner2023ACCESS}. We set the batch size as  64, 32, and 16 for Phi1.5-1.3b, Pythia 2.8, and Llama 7b, respectively. For Phi and Pythia, we evaluate on the test set every 3,000 training examples; for Llama, every 1,000 examples. All models are only learned for one epoch to avoid overfitting. 
 Algorithm \ref{alg} shows the pseudocode of alignment using subdata.

\RestyleAlgo{ruled}\label{alg}
\begin{algorithm}[h!]
\caption{\textit{Model Alignment Experiment}}\label{alg:align}
\SetKwInOut{Input}{Input}\SetKwInOut{Output}{Output}
\Input{Sub-data $\mathcal{D}_{\text{sub}}$ (random, hard, easy, or centroid), Base data $\mathcal{D}_{SFT}$, base language model $\pi_{0}$}
\Output{Optimal policy $\pi_{\theta}$}
\nl Obtain $\pi_{\text{ref}}$ via SFT of the base model $\pi_{0}$ using chosen pairs $(x,y^+)\in \mathcal{D}_{SFT}$\;
\nl Using $\pi_{\text{ref}}$ as the reference policy, align the model with DPO to obtain $\pi_{\theta}$ using $(x,y^+,y^-)\in \mathcal{D}_{aug}$ \;
\end{algorithm}

\subsection{Other loss function}
We explore the embedding-based method on other direct alignment methods such as SLiC and IPO. While this method is not proofed that be needed in the restriction of $\mathcal{D}\sim p^{*}$, we show the preliminary result in the Table for the HH-Helpful dataset, we found that dissimilar pairs also benefit from other direct alignment method
\begin{table}[h!]
\centering
  \caption{Learning Margins on SHP2 dataset.}
\begin{tabular}{cccc}
\Xhline{2\arrayrulewidth}
& {\textbf{Easy}}&{\textbf{Random}}&{\textbf{Hard}}\\
 \hline
SLiC  &\textbf{1.48}&1.33&0.92\\
IPO  &\textbf{1.43}&1.35&0.96\\
\Xhline{2\arrayrulewidth}
\end{tabular}
\label{tab:other loss}
\end{table}
\subsection{Easy bring lower reward}
\begin{figure*}
     \centering
     \begin{subfigure}[t]{0.32\textwidth}
         \centering        \includegraphics[width=\textwidth]{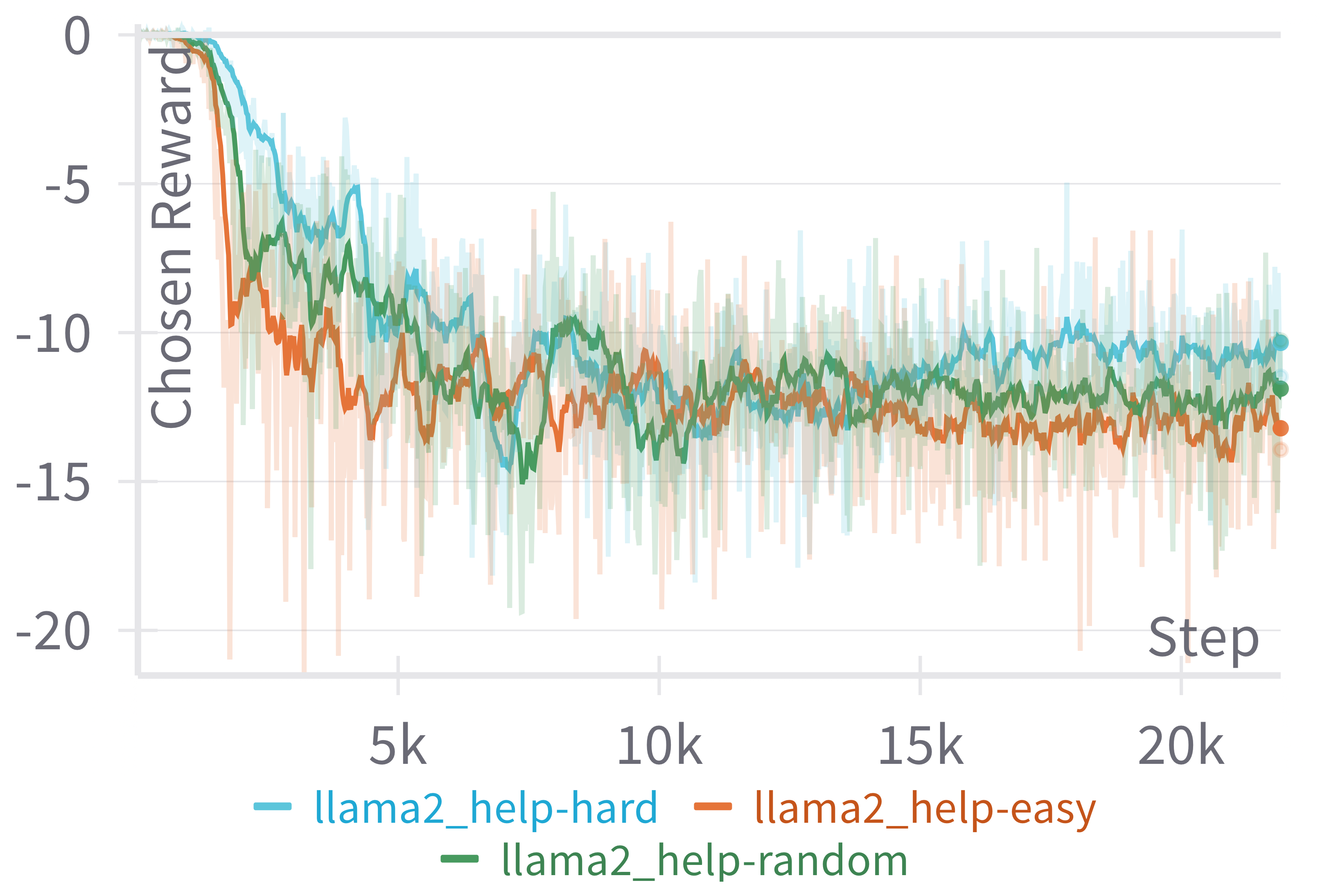}
         \caption{The chosen reward $r(y^+|x)$}
         \label{fig:train_chosen_r}
     \end{subfigure}
     \begin{subfigure}[t]{0.32\textwidth}
         \centering
         \includegraphics[width=\textwidth]{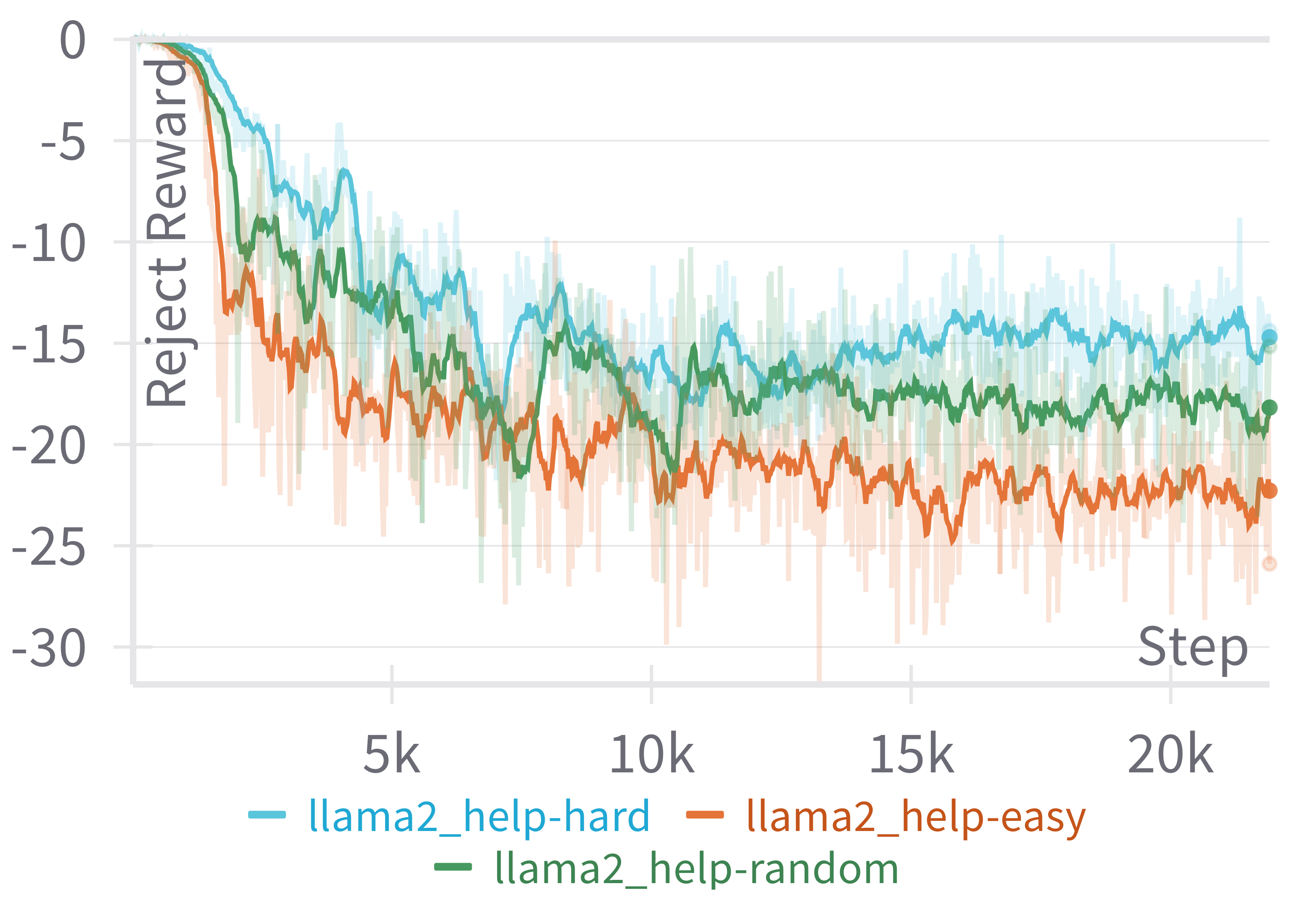}
         \caption{The rejected reward $r(y^+|x)$}
         \label{fig:train_reject_r}
     \end{subfigure}
     \begin{subfigure}[t]{0.32\textwidth}
         \centering
         \includegraphics[width=\textwidth]{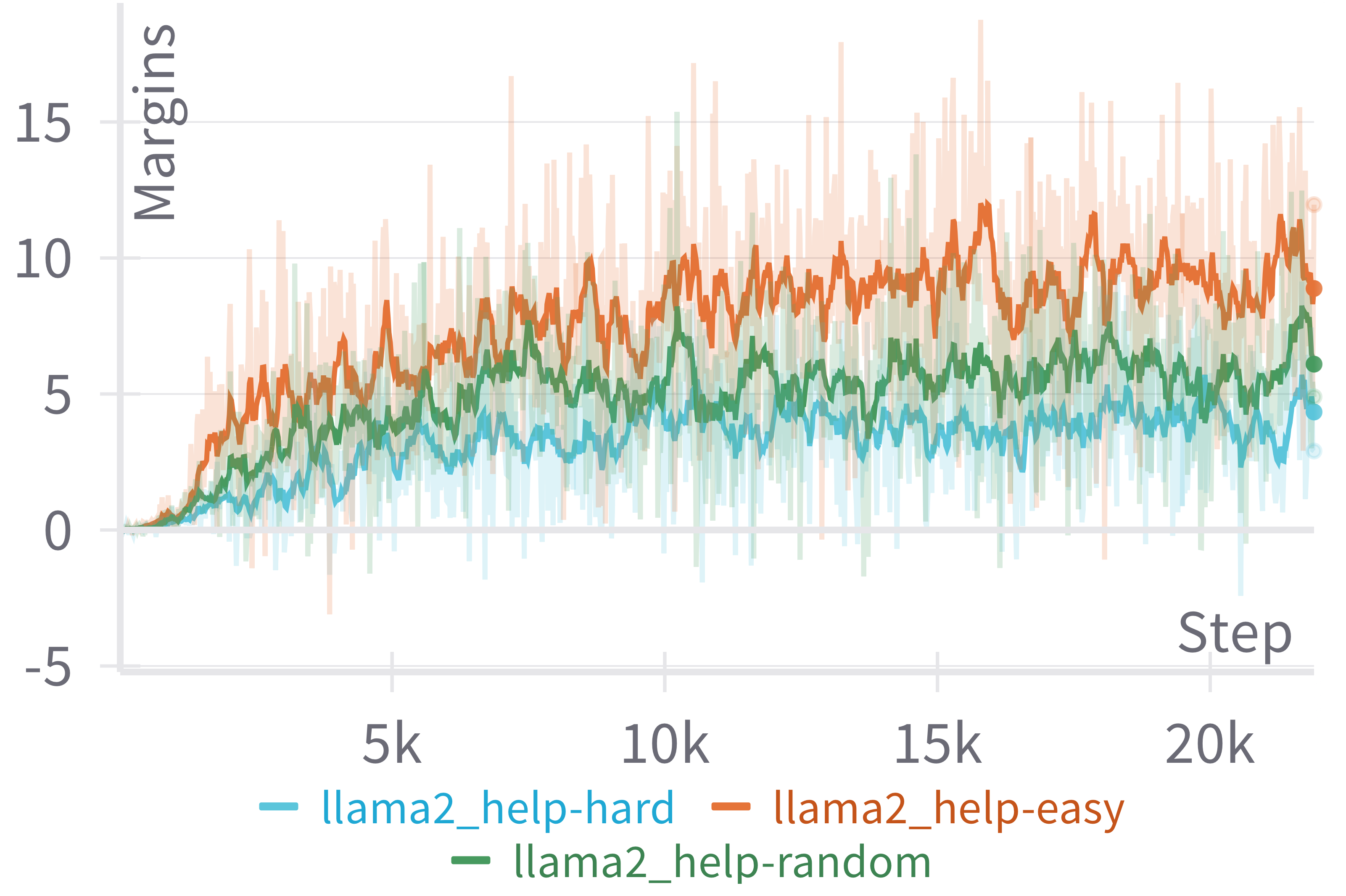}
         \caption{Training Margins}
         \label{fig:train_margin}
     \end{subfigure}
        \caption{Result of training on the different models}
        \label{fig:train_process}
\end{figure*}
We also compare the training process of different sub-sets, as shown in Fig.~\ref{fig:train_process}. From Figure \ref{fig:train_chosen_r} and Figure \ref{fig:train_reject_r}, the Easy will converge to a lower reward than the Hard one so it is more sensitive to the margins $r(x,y^+)-r(x,y^-)$ change. Slight changes in the reward for the ``\textbf{easy}'' will bring more margins than  ``\textbf{hard}'' and  ``\textbf{random}''. The ``\textbf{hard}''  has the larger reward on both chosen and rejected so will induce a smaller increase on the margins.
Therefore, the Easy obtains better margins in the training dataset, as shown in Fig. \ref{fig:train_margin}. 

Another explanation for the observed results might be that the prompts corresponding to the Easy response pairs have benign statistical qualities as, by construction of the subsets, the prompts are not shared between Easy and Hard subsets. To rule out this possibility, we conduct the experiments on another dataset, SHP2, where a single prompt corresponds to multiple responses. 

\section{Appendix E}\label{app:D}

\subsection{Examples of Sentence for calculate the embedding}\label{sentence example}
This is an example of the prompt and sentence for calculating the embedding in Fig.~\ref{fig:embedding change} 
\begin{mdframed}
    \texttt{\textbf{Human:}} Do you also happen to get sad because you simply will never have enough time to get the knowledge you want to? I was wondering if other people also get kind of sad because it is simply impossible to have enough time to learn everything you actually wanted to learn by a certain age/stage in life. Like idk, completing that list of books with important authors in your field, which gets longer and longer while keeping up with the recent findings in your field. Or learning an additional programming language or further practicing the one your familiar with. Or learning one more language. And all of that on top of the workload you already have.   Sometimes that makes me really sad because there are just so many things I am interested in and curious about but the more I study the things the longer the list of things I want to learn more about gets. Idk if you can relate but I just wanted to share this and I would be really interested to see what you think about this!

    \texttt{\textbf{Assistant:}} All the time. If I believed in an afterlife, I’d want it to basically be infinite time in the library.
\end{mdframed}

\subsection{Evaluator Prompt}\label{app:eva_prompt}
\begin{mdframed}
    \texttt{\textbf{System:}} You are a highly efficient assistant who evaluates and selects the best large language model (LLMs) based on the accuracy and harmlessness of their responses to a given instruction. This process will be used to create a leaderboard reflecting the most accurate and human-preferred answers.
\end{mdframed}

\begin{mdframed}
    \texttt{\textbf{User:}} I'll provide you with prompts given to these models and their corresponding outputs. Your task is to assess these responses and select the model that produces the best output from a human perspective.

\texttt{\textbf{Instruction:}}
\begin{verbatim}
{
    ``instruction'': ``{instruction}'',
}
\end{verbatim}

\texttt{\textbf{Model Outputs:}}
\texttt{Here are the unordered outputs from the models. Each output is associated with a specific model, identified by a unique model identifier.}

\begin{verbatim}
{
    {
        ``model_identifier'': ``m'',
        ``output'': ``{output_1}''
    },
    {
        ``model_identifier'': ``M'',
        ``output'': ``{output_2}''
    }
}
\end{verbatim}

\texttt{\textbf{Task:}}
Evaluate the models based on the quality and relevance of their outputs, and select the model that generated the best output. Answer by providing the model identifier of the best model. We will use your output as the name of the best model, so make sure your output only contains one of the following model identifiers and nothing else (no quotes, no spaces, no new lines, ...): m or M.
\end{mdframed}

\subsection{Response Example From different Data selecting Methods}

\begin{table*}[h!]
\centering
  \caption{Example of Responses.}
\begin{tabular}{p{1.6cm}||p{11cm}}
 \hline
 \textbf{Prompt} &\textcolor{blue}{Human: How do I understand/perceive non functional harmony? I'm a self taught musician and I'm trying to learn as much theory as possible. But I just can't seem to wrap my head around this topic. Nor am I able to understand how someone could ``feel'' this kinda music. If that makes sense that is. }\\
 \hline
\hline
 {Reference}&I think it's a matter of perception. If you're a trained musician, you're used to hearing music that sounds like it's not supposed to. You're used to hearing music that's not supposed to sound like it's supposed to.  \\
  \hline
Centroid & I think the best way to understand non functional harmony is to think of it as a kind of ``noise'' that is created by the instruments. It's not a ``harmonious'' sound, but it's not a ``bad'' sound\\
 \hline
{Easy}& I think it's a very interesting topic. I think it's important to understand that non functional harmony is not about the music itself, but about the way it makes you feel. It's about the emotional response that the music evokes in you.\\
 \hline
Hard & I think you're missing the point of non functional harmony. It's not about feeling the music, it's about understanding the music. It's about understanding the structure of the music, the way the notes are arranged, the way they interact with each\\
 \hline
Random & I think you're missing the point of non functional harmony. It's not about feeling it, it's about understanding it. It's about understanding the structure of the music and how it works. It's about understanding the relationships between the different parts of\\
\hline
\end{tabular}
\label{tab:generative result}
\end{table*}

\end{document}